\documentclass[letterpaper, 10 pt, conference]{ieeeconf}  

\IEEEoverridecommandlockouts                              

\usepackage{graphicx}

\usepackage{amsmath} 
\usepackage{amsfonts}
\usepackage{mathrsfs}
\usepackage{wrapfig}
\usepackage{cite}
\usepackage{graphicx}
\usepackage{textcomp}
\usepackage{xcolor}
\usepackage[capitalise]{cleveref}
\usepackage[list=on,listformat=simple]{subcaption}
\usepackage{preambles}
\usepackage{algorithm} 
\usepackage{algpseudocode}

\title{\LARGE \bf
Data-Driven Latent Space Representation for Robust Bipedal Locomotion Learning}
\author{Guillermo A. Castillo$^{1}$,  Bowen Weng$^{1}$, Wei Zhang$^{2}$, and Ayonga Hereid$^{3}$
\thanks{This work was supported in part by the National Science Foundation
under grant FRR-21441568.}%
\thanks{$^{1}$Electrical and Computer Engineering, Ohio State University, Columbus, OH, USA;  {\tt\footnotesize \{castillomartinez.2,weng.172\}@osu.edu.}}
\thanks{$^{2}$SUSTech Institute of Robotics, Southern University of Science and Technology (SUSTech), China; {\tt\footnotesize zhangw3@sustech.edu.cn.}}
\thanks{$^{3}$Mechanical and Aerospace Engineering, Ohio State University, Columbus, OH, USA. {\tt\footnotesize hereid.1@osu.edu.}}%
}

\begin{document}

\maketitle


\begin{abstract}


This paper presents a novel framework for learning robust bipedal walking by combining a data-driven state representation with a Reinforcement Learning (RL) based locomotion policy. The framework utilizes an autoencoder to learn a low-dimensional latent space that captures the complex dynamics of bipedal locomotion from existing locomotion data. This reduced dimensional state representation is then used as states for training a robust RL-based gait policy, eliminating the need for heuristic state selections or the use of template models for gait planning. The results demonstrate that the learned latent variables are disentangled and directly correspond to different gaits or speeds, such as moving forward, backward, or walking in place. Compared to traditional template model-based approaches, our framework exhibits superior performance and robustness in simulation. The trained policy effectively tracks a wide range of walking speeds and demonstrates good generalization capabilities to unseen scenarios.
\end{abstract}


\section{Introduction}\label{sec:intro}

Bipedal robots have long been a subject of fascination and research within the field of robotics due to their potential for versatile and agile locomotion in complex environments, closely mimicking the morphology and capabilities of humans.
However, despite significant advancements in control techniques and hardware, developing a robust and efficient bipedal locomotion control system remains a challenge, largely due to the high dimensionality, underactuation, and highly nonlinear and hybrid dynamics of bipedal locomotion. 

Conventional methods for bipedal walking often involve solving optimization problems using the robot's full-order~\cite{ames2013human} or reduced-order model~\cite{kajita2003biped,wensing2013high} to find feasible trajectories that enable stable walking gaits. 
The full-order model captures all the complexities and details of the robot's dynamics, but it can be computationally demanding and not suitable for real-time control~\cite{hereid2018dynamic}. On the other hand, reduced-order template models (such as linear inverted pendulum (LIP) and its variants~\cite{kajita2003biped,gong2022zero,xiong2021global}) simplify the dynamics of the system, making it easier to plan trajectories for the robot's center of mass and end-effector. However, these reduced-order models often require strict constraints to account for the mismatch between the reduced and full-order states of the robot.

\begin{figure*}[!t]
\centering
\includegraphics[trim={0cm 0cm 0cm 0cm},clip,width=1\linewidth]{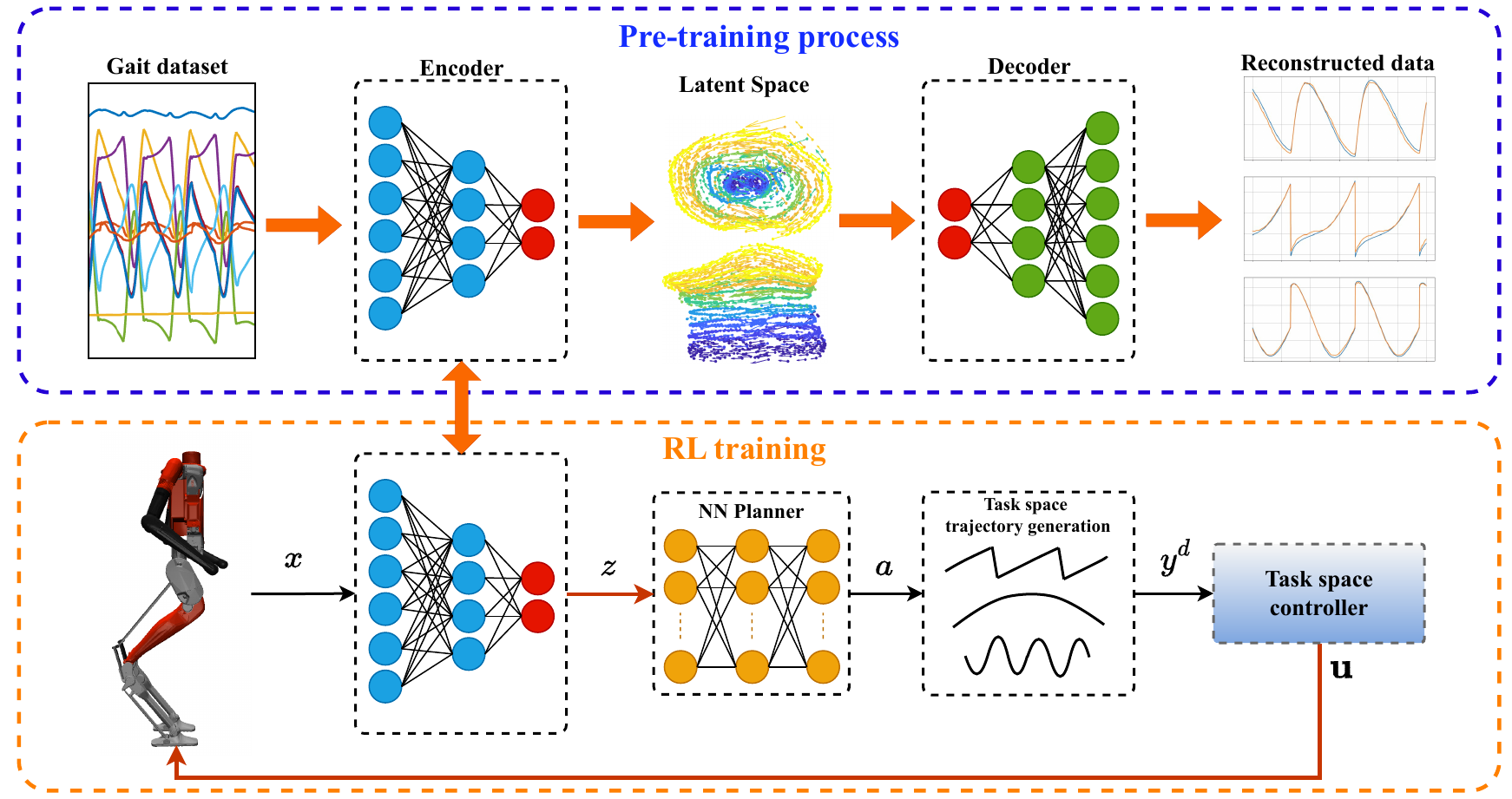}
\caption{An overview of the overall structure and flow of the proposed learning-based framework. In the pre-training phase, the autoencoder learns a latent space that captures the dynamics of the full-order system. During the RL training, the policy maps the latent representation to a set of task space actions that are translated into task space trajectories. Finally, a whole-body task space feedback controller computes the motor torque to track the desired task-space trajectories.}
\label{fig:overall_framework}
\vspace{-4mm}
\end{figure*}

The evolution of modern control theory has seen an infusion of machine learning and RL techniques, particularly with the growing abundance and accessibility of data. These data-driven approaches offer novel ways to address challenges in control design, often sidestepping traditional, more rigid, and computationally expensive methods. There is a growing interest in using reinforcement learning based approaches that allow exploiting data from simulation to train controllers in a model-free fashion~\cite{xie2018feedback, siekmann2020learning, li2021reinforcement, green2021learning}. However, similar to model-based techniques, they are highly dependent on the quality of the state representation provided to the learning algorithm. As an alternative, more complex frameworks have been proposed to combine learning algorithms with model-based controllers. In \cite{jimenez2022neural}, an HZD-based approach is used to learn a policy that satisfies Control Barrier Functions (CBF) defined on the reduced-order dynamics. In our previous work~\cite{castillo2021robust, castillo2022reinforcement}, a cascade structure is implemented to compensate the learned trajectories with feedback regulators to increase the robustness of the walking gait. A hierarchical structure that combines a template-based RL policy with a model-based low-level controller is proposed in \cite{castillo2023template}. 

An effective state representation that can accurately capture the complex dynamics of the whole system can significantly enhance the learning process, 
enabling more efficient learning and better transferability of control strategies.
Unsupervised learning, dimensionality reduction, and representation learning methods can be employed to extract relevant features from high-dimensional sensory data, enabling the development of more efficient and interpretable state representations~\cite{diederik217auto, bengio2013representation}.
Dai et al. \cite{dai2023datadriven} learn the step-to-step residual dynamics via an adaptive control approach, which is then used to design a foot-stepping controller for a bipedal robot in simulation. More complex learned residual dynamic models are also combined with Model Predictive Control (MPC) for agile systems \cite{salzmann2023real}. These techniques require previous knowledge of the dynamics of the system's model.
Several works also have exploited latent representations through reinforcement learning for locomotion. Peng et al. \cite{peng2022ase} combine techniques from adversarial imitation learning and unsupervised reinforcement learning to develop skill embeddings that produce locomotion behaviors. 
Starke et al. \cite{starke2022deepphase} extract a multi-dimensional phase space from the full-order state motion data, which effectively clusters animations and produces a manifold with better temporal and spatial alignment. However, these are end-to-end frameworks, which makes it difficult to establish a relationship between the latent space and the control actions of the policy. Moreover, these approaches have been mostly used to control animated characters in simulation, and there are no studies focused on implementation for actual bipedal robots.







In this paper, we propose a novel data-driven framework for bipedal walking that combines a learned low dimensional state representation of bipedal locomotion with a robust gait planner using RL, as depicted in \figref{fig:overall_framework}.  
Our framework uses an autoencoder to extract an effective reduced-dimensional state representation of the full-order system dynamics. We then integrate this reduced-order latent space with reinforcement learning and a task space feedback controller to train robust locomotion policies. 
This paper makes two key contributions. First, we demonstrate that the complex dynamics of bipedal robots can be effectively captured using a low-dimensional learned latent space. This allows for a more compact representation of the system's behavior. Second, we show that the learned latent variables can be leveraged to design a robust locomotion policy using RL. By bridging the gap between state representation learning and learning-based control policies, this work enables leveraging existing locomotion data to develop more effective and adaptable frameworks for versatile and robust bipedal locomotion.

\section{Preliminaries and Problem Formulation}\label{sec:problem_formulation}

\subsection{Hybrid System Model of Bipedal Locomotion}
The bipedal locomotion problem can be characterized as a hybrid system determined by a collection of phases of continuous dynamics with discrete events between the transitions of the continuous phases. Formally, the hybrid system model for biped locomotion can be defined as~\cite{grizzle2014models}:
\begin{align}\label{eq:hybrid_model}
  \Sigma: \left\{
\begin{array}{lcl}
    \dot{x}  = f(x) + g(x)u + \mathbf{\omega}(x,u) & \hspace{.01cm} & x \in \mathcal{X} \setminus \mathcal{H}\\
    x^+  = \Delta(x^-)       & \hspace{.01cm} & x^- \in \mathcal{H},
  \end{array}
\right.
\end{align}   
where $f(\cdot)$ and $g(\cdot)$ are vector fields, $x=(q,\dot{q}) \in \mathcal{X} \subseteq \R^n $ represents the robot states with $q$ being the vector of generalized coordinates, $u \in \mathcal{U} \subseteq \R^m$ is a vector of actuator inputs, and $\mathbf{\omega} \in \Omega \subseteq \R^w$ captures external disturbances and model uncertainties. 
The reset map $\Delta: \mathcal{H} \to \mathcal{X}$ denotes the mapping between the post-impact states $x^+$ immediately after impacts and the pre-impact states $x^-$ right before impacts, and $\mathcal{H}$ is the switching surface corresponding to swing foot impacts.

For a typical humanoid robot with high degrees of freedom (DoF), the dimension of the robot states $x$ is too large to be effectively used for feedback motion planning.
Low-dimensional models have become a powerful tool for motion planning of bipedal locomotion, given their potential to characterize the dynamics of bipedal walking into simple linear or nonlinear models. 
However, existing reduced-order template models, such as the Linear Inverted Pendulum (LIP), make assumptions that limit the full capabilities of walking robots. These assumptions include a constant center of mass (CoM) height and zero angular momentum about the CoM during the walking gait. While previous research has explored alternative template models for bipedal locomotion, this study aims to investigate whether available locomotion data can be utilized to identify an effective reduced-dimensional state representation of bipedal locomotion for motion planning purposes.

\vspace{-1mm}
\subsection{Data-driven Low Dimensional Latent Space}



Autoencoders are a great tool to harness high dimensional gait data to extract a reduced dimensional latent representation of the system that captures the essence of the full-order robot dynamics. An autoencoder works by compressing the input data into a low-dimensional latent space through a feature-extracting function in a specific parameterized closed form, such as neural networks~\cite{bengio2013representation}. This function, called \textit{encoder} is determined by 
\begin{align} \label{eq:encoder}
    z = h(x,\theta_e),
\end{align}
where $z$ is the latent variable or representation encoded from the input $x$, and $\theta_e$ is the vector of parameters for the encoder neural network.
Another closed-form parameterized function called the \textit{decoder}, maps from the latent space back to the full-order state. This function is defined by
\begin{align} \label{eq:decoder}
    \hat{x} = d(z,\theta_d),
\end{align}
where $z$ is the encoded latent variable, $\hat{x}$ is the reconstruction of the original input data $x$, and $\theta_d$ is the vector of parameters for the decoder neural network.

The strength of autoencoders lies in their ability to preserve most of the crucial information from the original data in the latent representation, even though it is of much lower dimensionality. This is ensured by training the autoencoder to minimize the reconstruction loss $\mathcal{L}$, which measures the error between the original data and its reconstruction from the latent space. In summary, autoencoder training consists of finding a value of the parameter vectors $\theta_e, \theta_d$ that minimize the reconstruction error: 
\begin{align} \label{eq:error_loss}
    \mathcal{J}_{\mathrm{AE}}(\theta_e, \theta_d)=\sum_{x^{(i)} \in \mathbf{X}} \mathcal{L}\left(x^{(i)}, d\left(h\left(x^{(i)},\theta_e\right),\theta_d\right)\right), 
\end{align}
where $x^{(i)}$ is a training sample containing the vectors of position and velocity of the generalized coordinates, and $\mathbf{X}$ is the training set containing all the data samples. $h$ and $d$ are the encoding and decoding functions introduced in \eqref{eq:encoder} and \eqref{eq:decoder}. Once the autoencoder has been trained, the resulting latent representation can then be utilized as part of the state for the high-level planner policy, offering a compact yet expressive state space for learning and control.

\section{Method}\label{sec:method}

The proposed data-driven framework is shown in ~\figref{fig:overall_framework}. The gait policy will be learned via RL by utilizing an effective latent representation of the full-order robot's dynamics to a set of task space commands that generate desired trajectories for the robot. Then, a whole body task space controller (TSC) from~\cite{castillo2023template} is employed to accurately track the desired trajectories to realize stable locomotion.

The proposed framework is tested with the robot Digit, which is a 3D fully actuated bipedal robot with $30$ DoF and $20$ actuated joints built by the company Agility Robotics. Each leg has six actuated joints corresponding to the motors located on the robot’s hip, knee, and ankle and three passive joints corresponding to the robot’s tarsus, shin spring and heel spring joints. In addition, it has four actuated joints per arm corresponding to the shoulder and elbow joints. Since the spring joints are very stiff, we considered them as fixed joints in this work. Therefore, the vector of generalized coordinates for Digit is defined by
\begin{align}\label{eq:digit_generalized_vector}
    q = (p, q_\phi, q_j),
\end{align}
where $p=(p_x, p_y, p_z)$ is the position of the robot's base, $q_\phi=(q_x, q_y, q_z, q_w)$ is the quaternion representation of the robot's orientation, and $q_j = (q_1,\dots,q_{n_j})$ is the vector of the robot's joints with $n_j=24$. Therefore, $q \in SE(3)\times \mathbb{R}^{24} \subset \mathbb{R}^{31}$, $\dot{q} \in \mathbb{R}^{30}$, and $x \in \mathbb{R}^{61}$.




\subsection{Reduced Dimensional Latent Space Representation}
To collect the locomotion dataset to train the autoencoder, we use the hierarchical controller proposed in \cite{castillo2023template}. 
The dataset is collected by performing walking gaits at various velocities. Specifically, the velocities $v_x$ and $v_y$ are varied within the ranges of $[-0.5, 1.0]$ m/s and $[-0.2, 0.2]$ m/s, respectively, with a step size of $0.1$ m/s, resulting in a total of $16$ different walking gaits.
Each walking gait has a duration of $10$ seconds, with the data being collected at a frequency of $50$ Hz. Therefore, the complete locomotion dataset $\mathbf{X}$ consists of $40,000$ samples of the robot's full-order states, i.e., $\mathbf{X}=\{x^{(i)} | i\in[1,40000]\}$.

\noindent \textbf{\textit{Remark: }} 
Note that any locomotion controller could be used to collect the gait data. For instance, many commercial robots are equipped with proprietary controllers that could be used to collect this data even though they are a black box from the user's perspective. Moreover, we do not make any assumptions about the distribution of the gait data.

In this work, we use an encoder parameterized by a fully connected neural network with three hidden layers of $128$, $64$, and $32$ units, respectively, and ReLU activation functions. The input of the encoder is the robot's full order states.
However, since the inputs of the encoder need to be bounded, the absolute base position of the robot cannot be directly used. This is because the absolute position can grow unbounded as the robot moves in the sagittal or frontal plane.
To address this, the base position and orientation of the robot are transformed from the world frame to the stance foot frame, which ensures that the inputs to the encoder remain bounded and allows for effective learning of the latent representation of the robot's dynamics.



The autoencoder is trained using Adam optimizer \cite{kingma2015adam} with a learning rate of $0.001$ and a batch size $B$ of $128$. The reconstruction loss is computed with the mean squared error (MSE) between the original values of the gait dataset $x^{(i)} \in \mathbf{X}$ and their reconstructed values $\hat{x}^{(i)} \in \mathbf{\hat{X}}$, given as
\begin{align}\label{eq:MES_loss}
    \mathcal{L} = \frac{1}{B} \left( x^{(i)} - \hat{x}^{(i)} \right)^2.
\end{align}
The autoencoder is trained for 400 episodes in a 12-core CPU machine with an NVIDIA RTX 2080 GPU. The training takes about 10 minutes using the described locomotion dataset $\mathbf{X}$ and PyTorch. 



\begin{figure}[t]
    \begin{center} 
    \includegraphics[trim={1.3cm 0.5cm 1.3cm 0.5cm},clip,width=1\linewidth]  {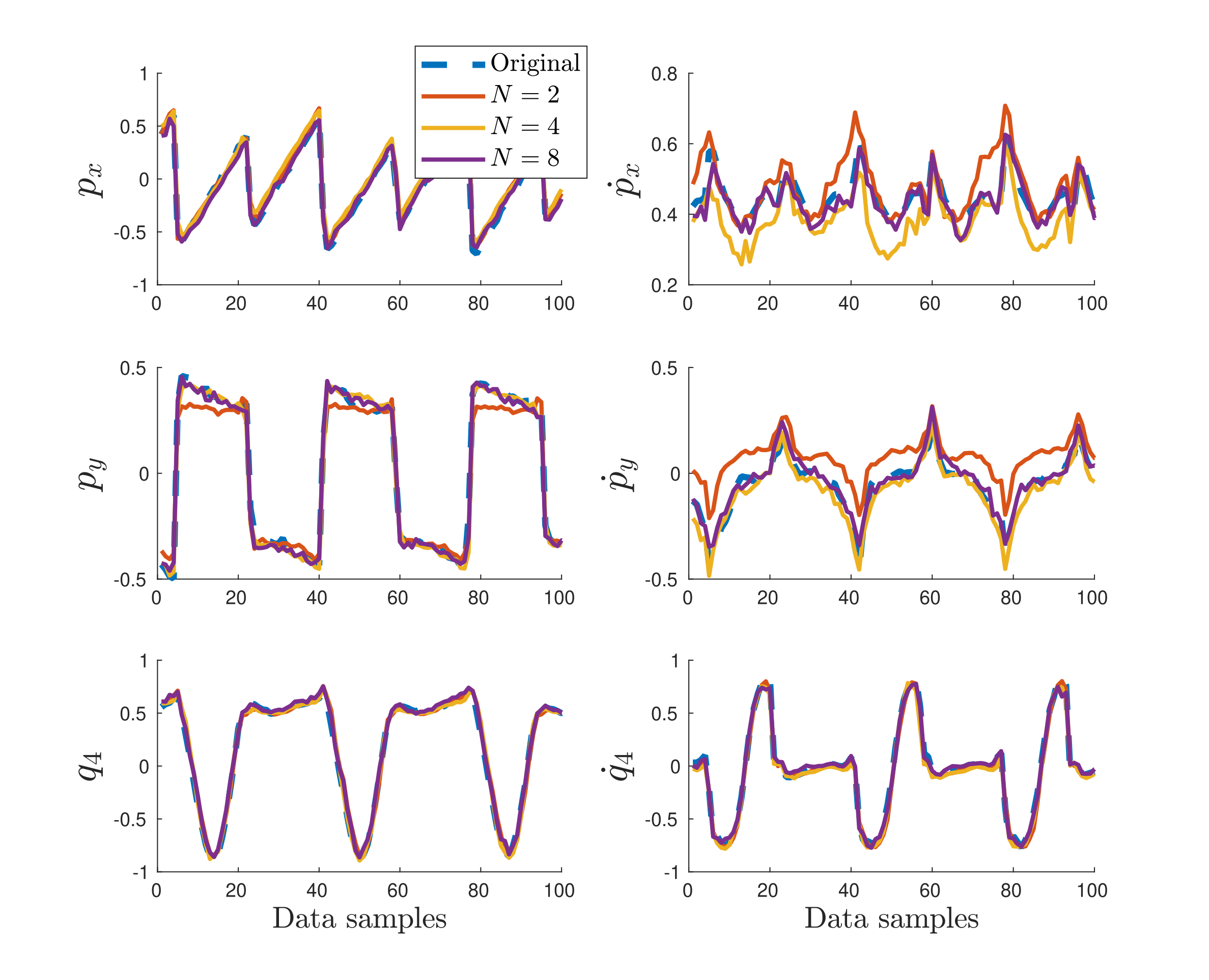} 
    \end{center} 
    \vspace{-2mm}
    \caption{Reconstruction of the robot's state with different dimensions $N$ for the latent variable $z$.  The plots show the position (left column) and velocity (right column) for the robot's base $x$ coordinate (top), robot's base $y$ coordinate (middle), and left knee joint (bottom), respectively.} 
    \label{fig:state_reconstruction}
    \vspace{-4mm}
\end{figure}

The selection of the dimension of the latent variable $z$ involves a trade-off. On one hand, a smaller dimension is desirable as it reduces the number of inputs for the RL policy. However, the dimension should also be large enough for the autoencoder to accurately reconstruct the full-order state of the system. In our study, we investigated this trade-off and discovered that even for a complex system like a humanoid robot, a latent variable dimension of $N=2$ is sufficient to capture the dynamics of the full-order systems effectively. This finding is supported by the results shown in \figref{fig:state_reconstruction}, where increasing the dimension of the latent variable does not lead to significant improvements in reconstruction quality. Even in cases where performance is slightly degraded, the reconstructed variable still follows the same pattern as the original state. 
Interestingly, our findings align with those reported in~\cite{starke2022deepphase}, where the authors demonstrated that periodic movements in bipedal locomotion can be represented using fewer than five phase variables. This similarity may be attributed to the symmetric and periodic nature of bipedal walking.

\subsection{RL-based Gait Policy using Learned Latent Variables} \label{subsec:rl_state}

Given the reduced dimensional latent space, we train an RL policy for robust locomotion based on our previous work in~\cite{castillo2023template}. The states of the policy are defined as 
\begin{align}
    s = (z, e_{\bar{v}}, v^d, a_{k-1}),
\end{align}
where $z=(z_1,\ldots, z_N)\in \mathbb{R}^N$ is the encoded latent state, $e_{\bar{v}}=(e_{\bar{v}_x},e_{\bar{v}_y})$ is the error between the average velocity, $\bar{v}=(\bar{v}_x,\bar{v}_y)$, and the desired velocity, $v^d = (v_x^d,v_x^d)$, of the robot, and $a_{k-1}$ is the last action of the planner policy.



\begin{figure}[t]
    \vspace{2mm}
    \begin{center} 
    \includegraphics[width=0.9\linewidth]{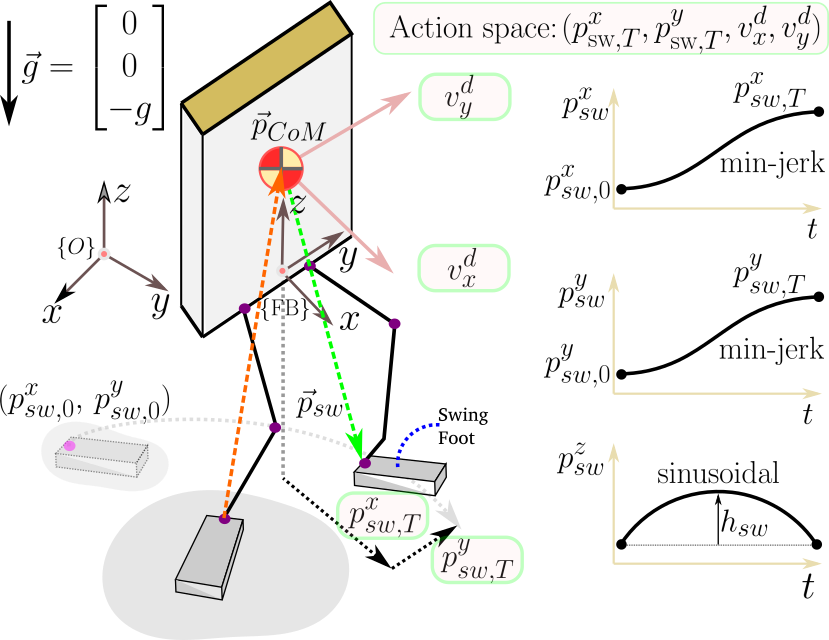} 
    \end{center} 
    \caption{Visualization of the policy actions and the trajectories generated for the task space controller. Additionally, we keep the torso straight up and the base at a constant height.} 
    \label{fig:robot_actions}
    \vspace{-5mm}
\end{figure}

The selection of the action space in this work is designed to exploit the natural nonlinear dynamics of the biped robot and enhance the robustness of the policy under various challenging scenarios.
Since Digit is a fully actuated system during the single support phase, we include the instantaneous base velocity as part of the action $a$. We choose to control instantaneous velocity over the base position because controlling the position during the stance phase through the TSC is more challenging than velocities in practical hardware implementation. This is caused by the noisy and inaccurate estimation of the position and the limited amount of torque in the robot's ankles. Moreover, sudden motions of the foot position due to terrain irregularities could produce big errors in the base position tracking that could result in aggressive control maneuvers. Therefore, controlling the instantaneous velocity is a less aggressive strategy for the TSC and provides some degree of damping to the ankle motion that helps with the stability of the walking gait under irregular terrains. 
Thus the action $a \in \mathcal{A}$ of the policy for Digit is chosen to be: 
\begin{align}
    a = (p^{x}_{\text{sw},T}, p^{y}_{\text{sw},T}, v_x^d, v_y^d), 
\end{align}
which corresponds to the landing position of the swing foot in $x$ and $y$ coordinates with respect to the robot's base and an offset to the instantaneous velocity of the robot's base in $x$ and $y$ coordinates, as illustrated in Fig. \ref{fig:robot_actions}.
The trajectory generation module transforms the policy action $a$ into smooth task-space trajectories for the robot's base and end-effectors. Specifically, the trajectory for the relative swing foot positions, $p^x_{\text{sw}}$ and $p^y_{\text{sw}}$, are generated using a minimum jerk trajectory connecting initial foot positions with target foot positions from the policy action. In particular, the initial foot positions will be computed at every touch-down event and kept constant throughout the current step.


The neural network (NN) chosen to parameterize the gait policy is a feed-forward network with two hidden layers, each layer with $128$ units. The hidden layers use the ReLU activation function, and the output layer is bounded by the $\mathrm{Tanh}$ activation function and a scaling factor to constrain the maximum value of the policy commands within the feasible physical limits of the robot hardware. Finally, we implemented a model-based TSC controller following the structure to track the task space trajectories, as described in our previous work~\cite{castillo2023template}. 



\begin{figure}[t]
    \begin{center} 
    \includegraphics[trim={2cm 0cm 0cm 0cm},clip,width=1\linewidth]  {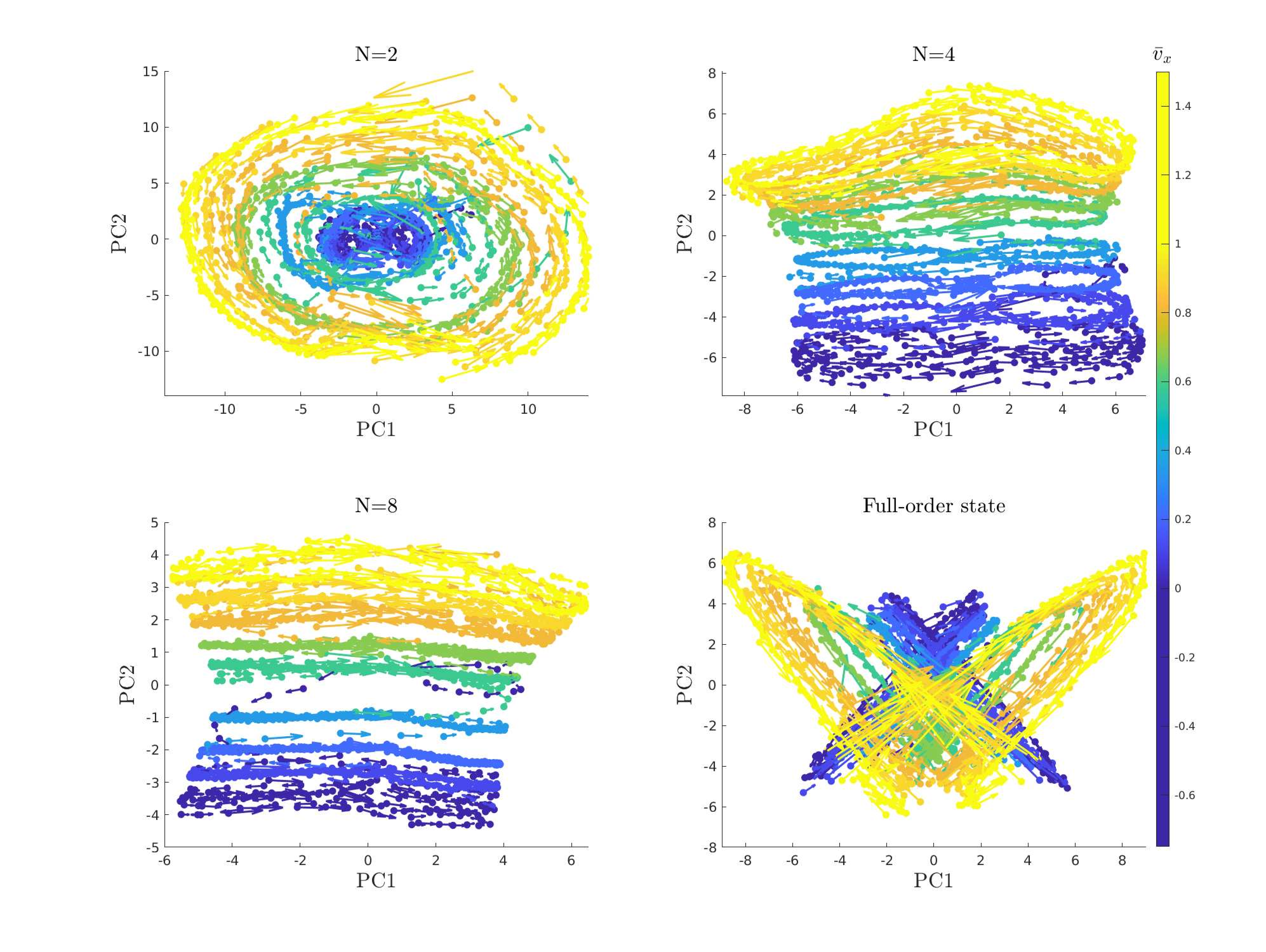} 
    \end{center} 
    \vspace{-5mm}
    \caption{Two-dimensional principal component analysis (PCA) of the learned latent manifolds at different walking speeds.} 
    \label{fig:latent_space}
    \vspace{-4mm}
\end{figure} 

\subsection{Learning Procedure of the RL Gait Policy} \label{learning_procedure}
To train the RL policy, we use the Proximal Policy Optimization algorithm~\cite{schulman2017proximal} with input normalization, fixed covariance, and parallel experience collection. We follow the algorithm implementation described in~\cite{siekmann2020learning}.

For each episode during the RL training, the initial state of the robot is set randomly from a normal distribution about an initial pose corresponding to the robot standing in the double support phase. 
An episode will be terminated early if the torso pitch and roll angles exceed $1$ rad or the base height falls below $0.8$ m.
The reward function adopted in this work is designed to (1) keep track of desired velocities, (2) minimize the angular momentum around the center of mass, denoted as $L_\textrm{CoM}$, and (3) reduce the variation of the policy actions between each iteration. More specifically,
\begin{align}
    \label{eq:reward}
    \mathbf{r} = \mathbf{w}^T [ r_{v},  r_{L_\textrm{CoM}}, r_{a}]^T,
\end{align}
with
\begin{align}
    \label{eq:reward_detailed}
    r_{v_x} &= \exp{({-\norm{\bar{v} - v^d}}^2)},\\
    r_{L_\textrm{CoM}} &= \exp{({-\norm{L_\textrm{CoM}}}^2)},\\
    r_{a} &= \exp{({-\norm{a_k - a_{k-1}}}^2)},
\end{align}
and the weights are chosen as $\mathbf{w}^T = [0.6, 0.3, 0.1]$.

One iteration step of the policy corresponds to the interaction of the learning agent with the environment. The RL policy takes the reduced order state $s$ and computes an action $a$ that is converted in desired task-space trajectories $y^d$ at the time $t_k$. The reference trajectories are then sent to the task-space controller, which sends torque commands to the robot. This workflow is depicted in \figref{fig:overall_framework}. The feedback control loop runs at a frequency of $1$ kHz, while the high-level planner policy runs at $50$ Hz. The maximum length of each episode is $600$ iteration steps, which corresponds to $12$ seconds of simulated time.

\vspace{2mm}
\section{Simulation results}\label{sec:results}
In this section, we show the performance of the learned planner policy under different testing scenarios with the bipedal robot Digit. Moreover, we analyze the latent representation generated during the testing of the RL policy.


\subsection{Latent State Representation} \label{sec:results_latent}


To visualize the learned latent manifolds, we applied principal component analysis (PCA) to reduce the dimensionality of the latent space to 2D. \figref{fig:latent_space} shows the 2D representation of the latent space for dimensions $N=2$, $N=4$, and $N=8$. The data used for visualization correspond to the robot Digit walking with the learned RL policy within the range of $v_x^d \in [-0.75,1.4]$ m/s. In all three cases, the latent space exhibits a well-distributed and disentangled representation of the data, with each walking speed corresponding to a specific area in the 2D plane. 
For comparison, \figref{fig:latent_space} also includes the 2D PCA visualization of the full-order state of the robot. It is evident that data points for different speeds overlap with no clear structure. This highlights that the latent space, even with a very low dimensionality $N$, effectively captures the distribution of the data in the full-order system. This demonstrates the potential of exploiting the latent representation for control purposes.

\begin{figure}[t]
    \begin{center} 
    \includegraphics[trim={0.5cm 0.0cm 1cm 0cm},clip,width=0.95\linewidth]  {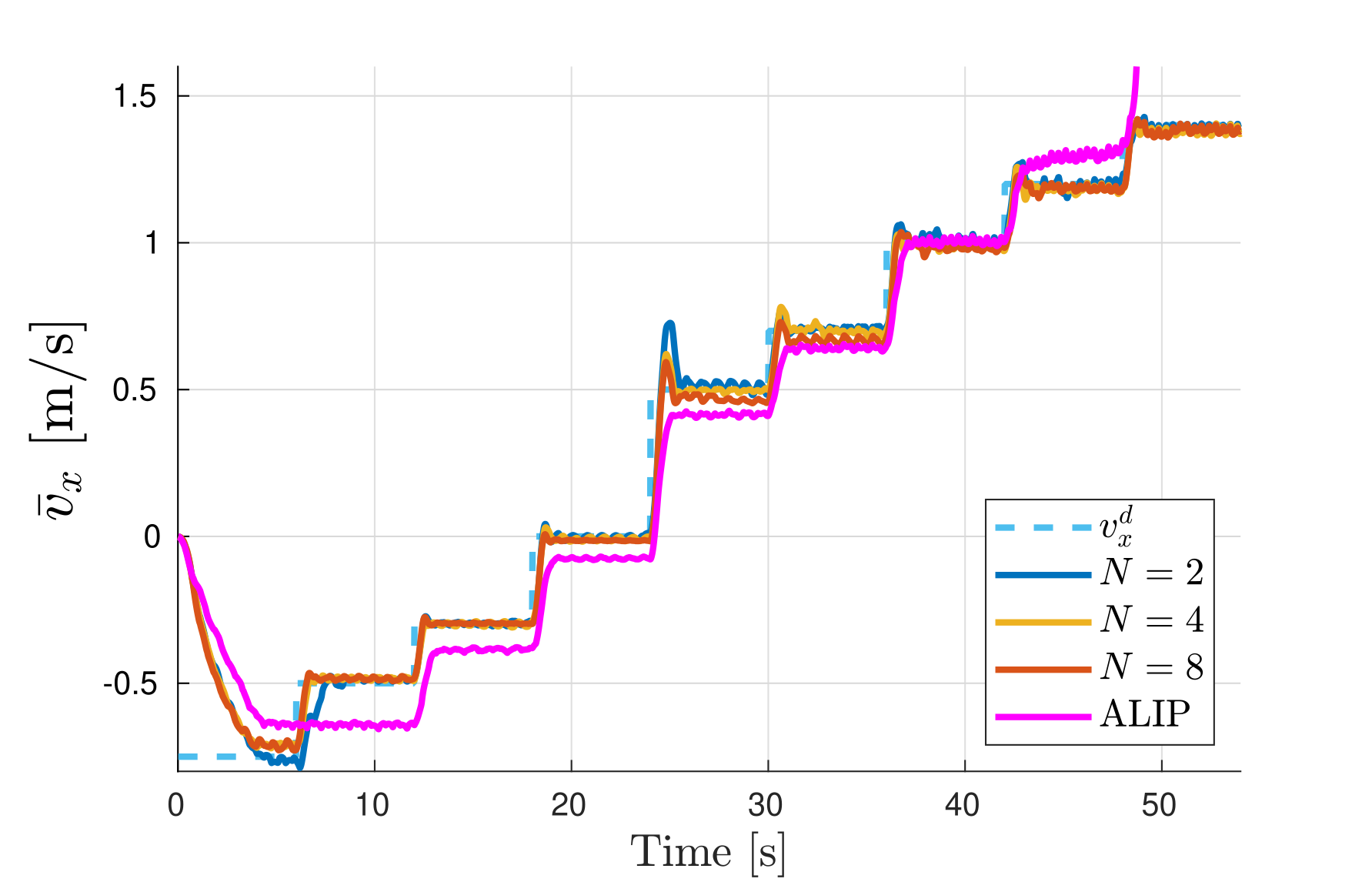}
    \end{center} 
    \vspace{-4mm}
    \caption{Comparison of velocity tracking performance of the learned policy with different dimensions of the latent space $N$ with a model-based ALIP planner described in~\cite{gong2022zero}.} 
    \vspace{-5mm}
    \label{fig:speed_tracking}
\end{figure}

\subsection{Tracking Performance of Different Velocity Profiles} \label{subsec:results_speed_tracking}
We evaluated the performance of the learned gait policies with latent space dimensions with $N=2$, $N=4$, and $N=8$ in tracking a velocity profile in different directions using the Digit robot. As shown in \figref{fig:speed_tracking}, the policy successfully tracks walking speeds in the range $v_x^d \in [-0.75,1.4]$ \ m/s, even with aggressive changes in the velocity profile. Importantly, the data collected to train the autoencoder was within the range of $v_x^d \in [-0.5,1.0]$.
These results show the effectivity of the learned latent states to control the walking velocity even with a low dimension $N$. 
Furthermore, the results demonstrate that the latent representation captures the dynamic nature of the walking gait and can be generalized to scenarios outside the training distribution.
Additional tests about this aspect are presented in Sec. \ref{subsec:generalization}.

\begin{figure}[t]
    \begin{center} 
    \includegraphics[trim={0.5cm 0.0cm 1cm 0cm},clip,width=1\linewidth]  {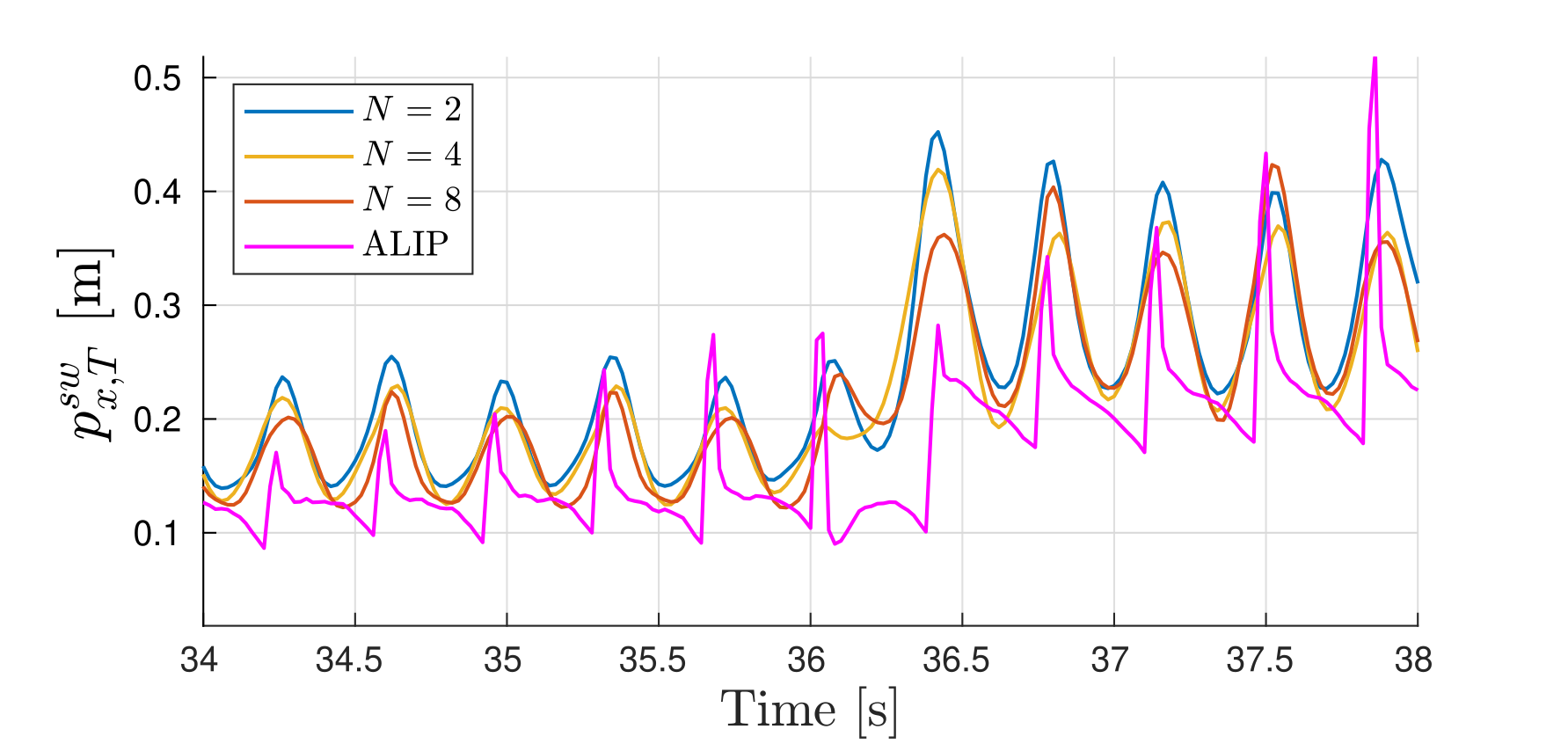}
    \end{center} 
    \vspace{-4mm}
    \caption{Comparison of the desired swing foot landing positions (the policy action) from different latent space dimensions $N$ and the ALIP planner.} 
    \label{fig:policy_actions}
    \vspace{-5mm}
\end{figure}

For comparison, \figref{fig:speed_tracking} also shows the speed-tracking performance of a model-based ALIP planner based on the design in \cite{gong2022zero}. 
Our policy outperforms the ALIP-based planner in terms of velocity tracking and offers a wider range of admissible walking speeds. The ALIP planner fails to maintain a stable walking gait for speeds higher than 1.2 m/s.
In addition, \figref{fig:policy_actions} illustrates the correspondence between the walking speeds and control actions of the RL policies with different $N$s and the ALIP planner. Specifically, we focus on the swing foot landing position with respect to the robot's base $p^{sw}_{x, T}$ when the robot is walking at $0.7$ and $1.0$ m/s. Interestingly, the actions of the RL policy exhibit similar patterns across different latent space dimensions $N$. This finding aligns with the results presented in Sec. \ref{sec:results_latent}, where we showed that even a small $N$ is sufficient to characterize the dynamics of the full-order system fully. 
When provided with a larger latent space dimension, the policy learns to disregard less important states in the latent space, resulting in similar policy actions for different $N$ values.

Furthermore, the actions of the latent space-based policies exhibit similar patterns to those of the ALIP planner. This is intriguing because the latent space used by the RL policy does not have a direct physical interpretation. Nevertheless, the RL policy learns to behave in a manner similar to the template model-based planner. These behaviors are not enforced during training, suggesting that the latent space naturally captures the dynamics of both the template model and the full-order system.

\subsection{Policy Generalization to Out-of-Distribution Scenarios} \label{subsec:generalization}

\begin{figure}[t]
    \begin{center} 
    \includegraphics[width=0.95\linewidth]{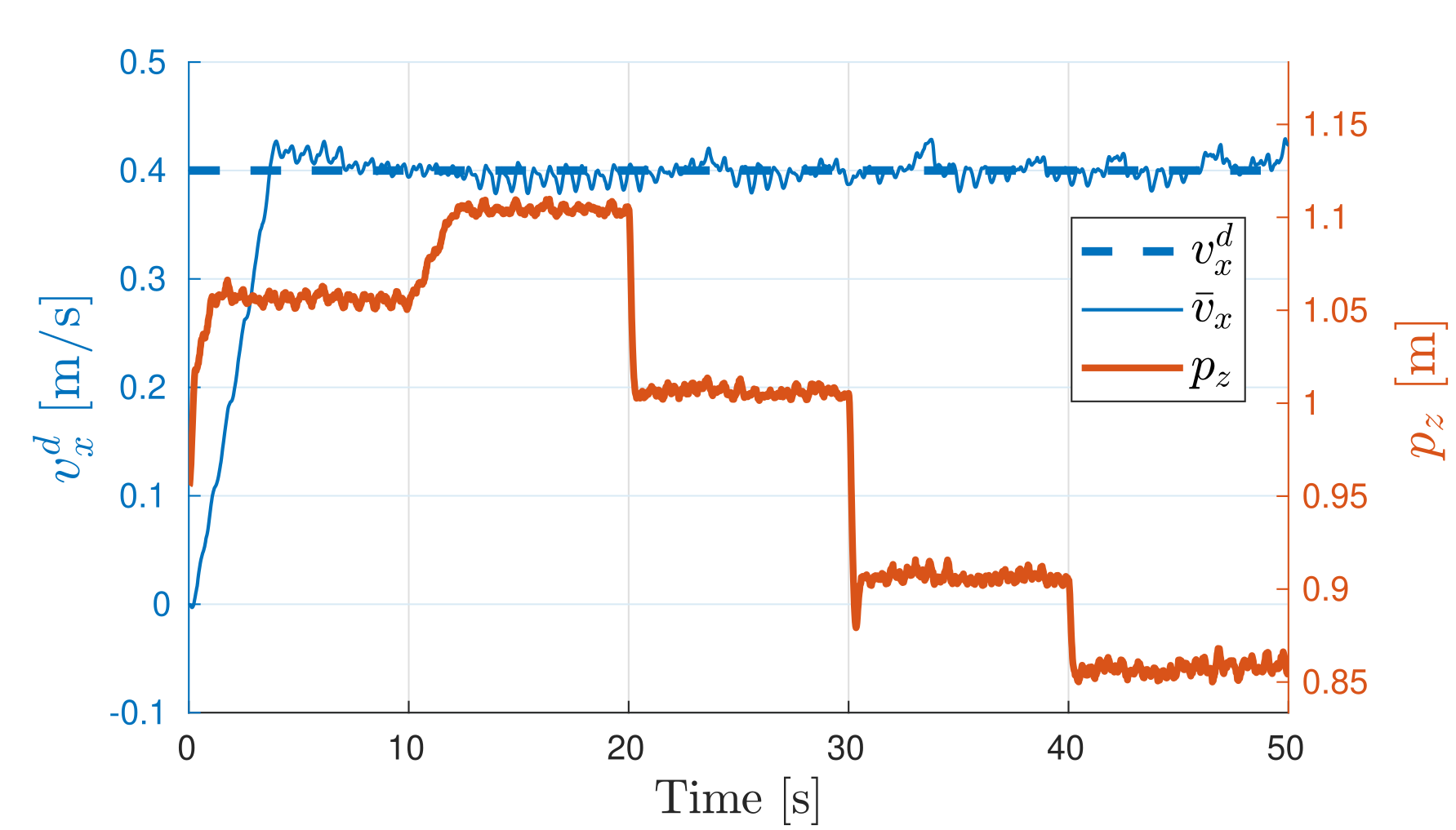}
    \end{center} 
    \vspace{-4mm}
    \caption{Generalization of the latent space and trained policy to out-of-distribution data.} 
    \label{fig:generalization}
\end{figure}

In addition to evaluating the policy's performance on data it was trained on, we also assess its ability to handle out-of-distribution scenarios. To do this, we conduct tests where the policy is instructed to maintain a constant speed while the height of the base is varied. It is important to note that the training of the autoencoder and the RL policy did not include any data with varying base heights, as the locomotion data used for training latent spaces was collected with a fixed base height of 1 m. However, as depicted in \figref{fig:generalization}, the policy demonstrates successful tracking of the desired walking speed regardless of the different base heights commanded.
An interesting future work direction may be the exploration of the robust generalization of the latent space to generate transitions between different locomotion tasks such as walking, jogging, sitting, and jumping. 

\begin{figure}[t]
    \begin{center} 
    \includegraphics[width=0.95\linewidth]{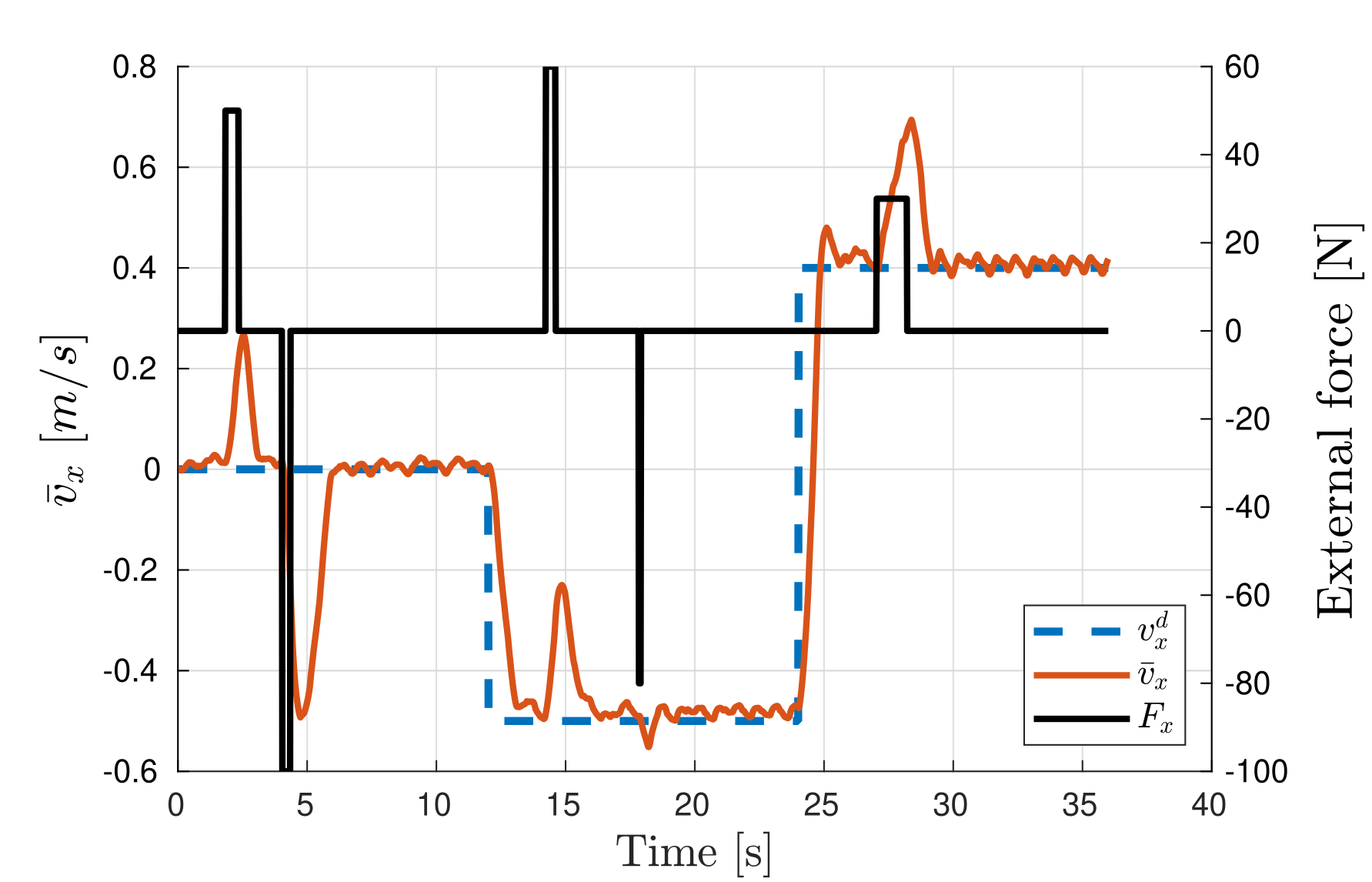}
    \end{center} 
    \vspace{-4mm}
    \caption{Robustness test to external disturbances.} 
    \vspace{-5mm}
    \label{fig:robustness_force}
\end{figure}

Furthermore, we conducted tests to evaluate the policy's robustness against external disturbances in the forward and backward directions. The disturbances ranged from $-100$ N to $60$ N, with durations ranging from $0.1$ s to $1.5$ s. As illustrated in \figref{fig:robustness_force}, the policy demonstrated effective reactions to the different disturbances, successfully maintaining stability and tracking the desired walking speeds without falling.

\section{Conclusion}\label{sec:conclusion}
In this work, we present a novel data-driven learning framework to realize robust bipedal locomotion.
The design of the high-level RL gait policy takes data-driven reduced dimensional latent variables as input states and generates a set of task space commands, including the robot's step length with respect to the base and instantaneous velocity offset of the robot's base. The latent representation of the full-order state is obtained using an autoencoder trained with supervised learning from locomotion data collected with existing locomotion controllers. Our work shows that the learned latent representation manifold has a disentangled structure that is directly correlated with the speed of the walking robot. The insightful choice of the RL state and action spaces results in a compact policy that learns effective strategies for robust and dynamic locomotion in simulation. 
Future work will focus on implementing and validating the proposed framework on Digit, further demonstrating its effectiveness and adaptability for real-world bipedal locomotion.





\clearpage
\bibliographystyle{IEEEtran}
\bibliography{references.bib}  

\end{document}